\documentclass[10pt, a4paper]{article}
\usepackage{booktabs}
\usepackage{listings}
\usepackage{multirow}
\usepackage{tcolorbox}
\usepackage{xcolor,soul}
\usepackage{graphicx}

\definecolor{lightyellow}{RGB}{255, 255, 180}
\definecolor{lightblue}{RGB}{180, 220, 255}
\definecolor{lightgreen}{RGB}{200, 255, 200}
\definecolor{lightred}{RGB}{255, 200, 200} 

\newcommand{\hlblue}[1]{{\sethlcolor{lightblue}\hl{#1}}}
\newcommand{\hlyellow}[1]{{\sethlcolor{lightyellow}\hl{#1}}}
\newcommand{\hlgreen}[1]{{\sethlcolor{lightgreen}\hl{#1}}}
\newcommand{\hlred}[1]{{\sethlcolor{lightred}\hl{#1}}}

\usepackage{tabularx, booktabs, multirow}
\usepackage[final]{lrec2026} 

\title{FAME: Fictional Actors for Multilingual Erasure}

\name{Claudio Savelli\textsuperscript{1}, Moreno La Quatra\textsuperscript{2}, Alkis Koudounas\textsuperscript{1}, Flavio Giobergia\textsuperscript{1}} 

\address{\textsuperscript{1}Politecnico di Torino, Italy \quad \textsuperscript{2}Università degli Studi di Enna "Kore", Italy \\
\textbf{Correspondence:} \href{mailto:claudio.savelli@polito.it}{claudio.savelli@polito.it}
}

\abstract{
LLMs trained on web-scale data raise concerns about privacy and the right to be forgotten.
To address these issues, Machine Unlearning provides techniques to remove specific information from trained models without retraining from scratch.
However, existing benchmarks for evaluating unlearning in LLMs face two major limitations: they focus only on English and support only entity-level forgetting (removing all information about a person).
We introduce FAME (Fictional Actors for Multilingual Erasure), a synthetic benchmark for evaluating Machine Unlearning across five languages: English, French, German, Italian, and Spanish.
FAME contains 1,000 fictional actor biographies and 20,000 question-answer pairs.
Each biography includes information on 20 topics organized into structured categories (biography, career, achievements, personal information).
This design enables both entity-level unlearning (i.e., forgetting entire identities) and instance-level unlearning (i.e., forgetting specific facts while retaining others). We provide two dataset splits to support these two different unlearning scenarios and enable systematic comparison of unlearning techniques across languages.
Since FAME uses entirely fictional data, it ensures that the information was never encountered during model pretraining, allowing for a controlled evaluation of unlearning methods.
 \\ \newline  }

\begin{document}

\maketitleabstract

\section{Introduction}

Large Language Models have transformed natural language processing by achieving state of the art performance across diverse tasks.
However, their training on massive web-scale datasets introduces significant concerns about data privacy and the \textit{``right to be forgotten''} \cite{mantelero2013eu}.
LLMs are known to memorize and reproduce sensitive information from their training data, including personal details, copyrighted content, and confidential information \cite{carlini2022quantifying,carlini2019secret}.
Privacy regulations, such as the European Union's General Data Protection Regulation (GDPR), grant individuals the right to request the deletion of their personal information from data systems \cite{juliussen2023algorithms}.
Retraining LLMs from scratch after removing specific data is prohibitively expensive, often costing millions of dollars and requiring weeks of computation \cite{hoffmann2022training}.
Machine Unlearning offers a practical alternative by selectively removing the influence of specific data from trained models without full retraining.

To develop and compare effective unlearning methods for LLMs, the research community requires robust evaluation benchmarks.
A fundamental challenge in creating such benchmarks is ensuring that evaluation data was not present in the model's pretraining corpus \cite{eldan2023s}.
Without this guarantee, it becomes impossible to determine whether the model truly forgot information or simply never learned it in the first place.
The opaque nature of most LLM training datasets makes it difficult to verify which information a model has previously encountered.
This challenge motivates the use of synthetic, fictional data that is guaranteed to be novel to the model.

TOFU \cite{maini2024tofu} introduced this approach by proposing 200 fictional author profiles for evaluating unlearning in LLMs.
While TOFU provided an important foundation, it has two key limitations that restrict its applicability to modern multilingual LLMs and real-world privacy scenarios.
First, TOFU focuses exclusively on English, despite the widespread deployment of multilingual models and the global nature of privacy regulations.
Recent work has shown that unlearning in one language does not necessarily transfer to others \cite{choi-etal-2024-cross}, underscoring the need for multilingual evaluation.
Second, TOFU supports only entity-level unlearning, where all information about a person must be removed simultaneously.
Real-world privacy requests, however, often involve partial information removal.
For example, an individual might request deletion of their contact information while allowing their professional achievements to remain accessible.
Current benchmarks cannot evaluate such fine-grained, instance-level unlearning scenarios.

We introduce FAME (Fictional Actors for Multilingual Erasure), a synthetic benchmark designed to address these limitations.
FAME contains 1,000 fictional actor biographies distributed across five languages: English, French, German, Italian, and Spanish.
Each biography is structured as exactly 20 atomic facts organized into four semantic categories: biography, career, achievements, and personal information.
This structured design enables controlled evaluation of both entity-level unlearning (removing all information about an actor) and instance-level unlearning (removing specific facts while retaining others).
The dataset includes 20,000 question-answer pairs, with each fact represented by a single QA pair, enabling precise measurement of which information has been forgotten.

FAME makes three key contributions.
First, it provides the first multilingual benchmark for evaluating MU in LLMs, enabling systematic analysis of how unlearning methods perform across different languages within the same model.
Second, it supports instance-level unlearning via its structured atomic facts, enabling the evaluation of partial-deletion scenarios that reflect real-world privacy requests.
Third, it offers two complementary dataset splits: an entity-based split for evaluating complete identity removal, and a topic-based split for evaluating selective fact deletion.
This dual-split design enables comparisons of unlearning methods across different forgetting scenarios.

We use FAME to evaluate five unlearning methods on a multilingual LLM across all five supported languages.
Our experiments assess unlearning methods across the three key dimensions that define the task \cite{hayes2024inexact, koudounas25c_interspeech}: \textit{utility}, measuring how well the model preserves its performance on retained knowledge; \textit{efficacy}, quantifying the degree to which the targeted information is successfully removed; and \textit{efficiency}, capturing the computational cost of achieving forgetting. Evaluating along these complementary axes enables a comprehensive comparison of approaches and highlights the trade-offs involved in multilingual and instance-level settings.
All data and evaluation code will be released publicly to support future research in Machine Unlearning.

\section{Related Work}
\label{sec:related_work}

Machine Unlearning aims to remove the influence of specific data from a trained model so that it behaves as if such data had never been seen.
While initially explored in the context of classification and vision tasks \cite{bourtoule2021machine}, adapting these techniques to LLMs has become increasingly important due to their tendency to memorize training data \cite{carlini2022quantifying,carlini2019secret} and the prohibitive computational costs of retraining from scratch \cite{hoffmann2022training, Crawford2022}.

\subsection{Machine Unlearning Methods for LLMs}

Recent work has demonstrated that various unlearning methods can be applied to pre-trained language models with significantly greater efficiency than full retraining.
\citet{yao-etal-2024-machine} show that unlearning approaches can be over $10^5$ times more computationally efficient than retraining while addressing diverse objectives, including removing harmful content and erasing copyright-protected information.
\citet{liu2025rethinking} provide a systematic review of unlearning methodologies, covering gradient-based approaches, optimization-based methods, and in-context unlearning techniques.
These methods vary in their approach to balancing forgetting efficacy with the preservation of model utility on retained knowledge, a trade-off that remains central to unlearning research.

\subsection{Benchmarks for Evaluating LLM Unlearning}

To evaluate unlearning techniques, several benchmarks have been proposed.
\citet{maini2024tofu} introduced TOFU, a controlled benchmark based on 200 synthetic author profiles, each represented by a biography and 20 question-answer pairs.
Since these fictional identities do not appear in any pre-training corpus, TOFU enables precise evaluation of whether a model forgets information introduced only during fine-tuning.
The benchmark measures forgetting efficacy and utility preservation across multiple unlearning methods, revealing that none of the tested algorithms achieved complete deletion without degrading model performance.
However, TOFU is limited to English and supports only entity-level forgetting, requiring the simultaneous removal of all information about a person.

Other recent benchmarks have addressed specific aspects of unlearning evaluation.
\citet{shi2024muse} proposed MUSE, a comprehensive framework that assess unlearning effectiveness across six criteria, including the removal of verbatim and semantic memorization, prevention of privacy leakage and preservation of model capabilities.
MUSE focuses on removing large text segments and demonstrates that most methods only partially mitigate memorization while significantly degrading performance on retained data.
\citet{hu2025blur} introduced BLUR to address realistic scenarios in which the forget and retain sets may overlap, arguing that previous benchmarks provide overly optimistic assessments by assuming a clean separation between what should be forgotten and what should be retained. 
\citet{wang2025towards} propose specific refinements to evaluation protocols that better capture the robustness of unlearning methods and the trade-offs between forgetting and retention.

Despite these advances, \citet{thaker2025position} argue that existing benchmarks provide weak measures of progress, as they are vulnerable to modifications that introduce dependencies between forget and retain information and may not reflect real-world unlearning scenarios.
All of these benchmarks, however, share a common limitation: they focus exclusively on English-language data.

\subsection{Entity-Level versus Instance-Level Unlearning}

A critical distinction in unlearning research concerns the granularity of forgetting.
\citet{ma-etal-2025-unveiling} formalize entity-level unlearning as the task of erasing all knowledge related to a specific entity from a model's parameters.
This approach reflects scenarios in which an individual requests the complete removal of their information, as might be required under privacy regulations.
In contrast, instance-level unlearning targets specific facts or attributes while preserving other information about the same entity.
These two paradigms require different evaluation strategies and pose distinct technical challenges~\cite{ma-etal-2025-unveiling}.
\citet{choi-etal-2025-opt} explore entity-level unlearning with the goal of erasing all entity-related knowledge while preserving the model's general capabilities.

Most existing benchmarks, including TOFU, focus primarily on entity-level unlearning due to the difficulty of generating controlled question-answer pairs that isolate individual facts without overlap.
This limitation prevents systematic evaluation of partial deletion scenarios, which are common in real-world applications where users may request removal of specific information (e.g., contact details or employment history) while allowing other facts to remain.

\subsection{Multilingual Evaluation}

The importance of multilingual evaluation has been widely recognized in natural language processing.
\citet{ahuja-etal-2022-beyond} highlight that most benchmarks cover only a handful of languages with limited linguistic diversity, restricting our understanding of model capabilities across languages.
\citet{huang2025benchmax} demonstrate, through large-scale evaluation, that the language-agnostic capabilities of LLMs remain uneven, with performance varying significantly across languages, even in multilingual models. 
This observation suggests that unlearning techniques effective in English may not generalize equally well to other languages.

Recent work has begun to address multilingual aspects of Machine Unlearning.
\citet{choi-etal-2024-cross} demonstrate that unlearning in one language does not necessarily transfer to others in multilingual models, making them vulnerable to low-resource language attacks where sensitive information remains accessible in less dominant languages.
They propose an adaptive unlearning scheme with language-dependent weights to address this cross-lingual challenge.
\citet{koudounas25c_interspeech} introduce UnSLU-BENCH, a benchmark for Machine Unlearning in spoken language understanding across four languages, focusing on speaker-level data removal.
Their findings reveal significant differences in the effectiveness of unlearning techniques across languages.

Despite the availability of multilingual LLMs such as BLOOM \cite{workshop2022bloom} and Llama \cite{touvron2023llama}, existing text-based unlearning benchmarks do not support evaluation across multiple languages.
This gap is particularly problematic given that privacy regulations such as GDPR \cite{juliussen2023algorithms,sartor2020impact} apply across multiple European languages. Yet, we lack tools to verify whether unlearning methods work consistently across linguistic boundaries.

\subsection{Positioning FAME}

Our work addresses these limitations by introducing FAME, a synthetic benchmark that enables evaluation of Machine Unlearning across five languages (English, French, German, Italian, and Spanish) and supports both entity-level and instance-level unlearning scenarios.
By structuring each biography into exactly 20 atomic facts organized into well-defined topics, FAME enables controlled evaluation of partial deletions while keeping the advantages of synthetic data used in TOFU.
The dual-split design (entity-based and topic-based) allows researchers to systematically compare unlearning methods across both complete entity removal and selective fact-forgetting conditions.

\section{Dataset Construction}

\begin{figure*}
    \centering
    \includegraphics[width=0.95\linewidth]{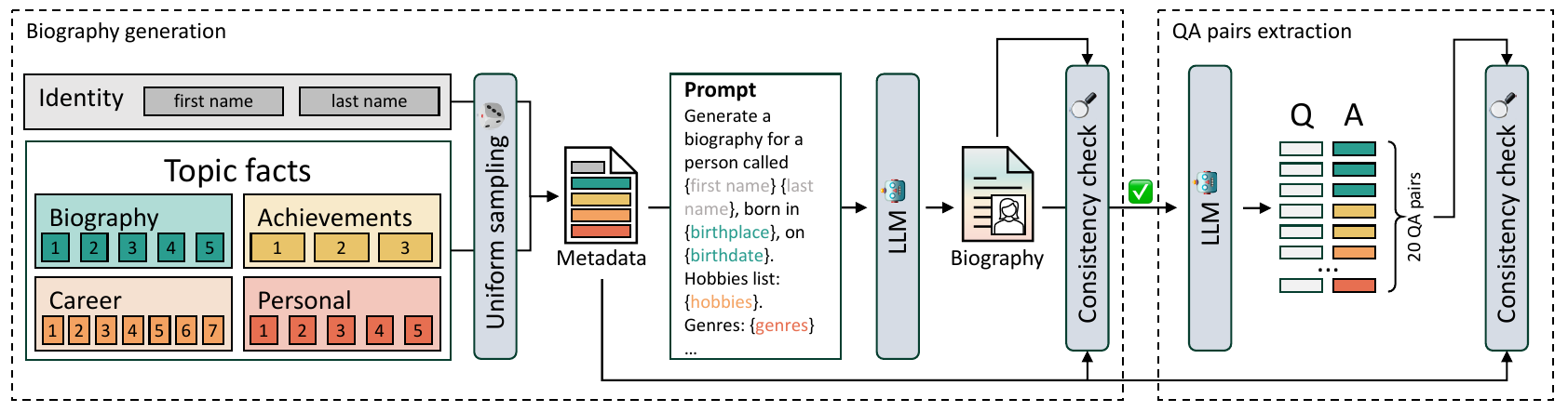}
    \caption{Overview of the FAME dataset generation pipeline. 
The process involves two main stages: (i) biography generation, where a large language model produces structured fictional profiles from predefined metadata and prompts; and (ii) question-answer extraction, where each atomic fact is converted into a QA pair. All outputs undergo validation procedure ensuring syntactic correctness and semantic alignment.}
    \label{fig:generation}
\end{figure*}
This section presents the pipeline adopted to create the dataset, summarized in Figure \ref{fig:generation}. The process includes an initial \textit{biography generation} step, followed by the \textit{extraction of question-answer pairs}, as detailed in Section \ref{ssec:definition}.  
Subsequently, Section~\ref{sec:split_statistics} introduces the two complementary dataset splits and provides a quantitative overview of the structure and composition of FAME, while Section~\ref{sec:diversity} describes the constrained generation strategy adopted to enhance dataset diversity and prevent repetitive or stereotypical patterns.

\subsection{Dataset Definition}
\label{ssec:definition}
Within the dataset, each instance represents a fictional actor or film director described through a structured biography and an associated set of question-answer pairs, in one of five languages: English, French, German, Italian, and Spanish. The dataset is entirely generated through a reproducible two-stage pipeline:
\textit{biography generation} and \textit{question-answer extraction}. Each phase is controlled by a set of deterministic prompts and verified through schema-based validation to ensure factual consistency and completeness.

\paragraph{Biography generation.}
For each target language, the pipeline constructs a unique fictional identity defined by structured metadata: name, date and place of birth, profession, hobbies, education, filmography, awards, and residence, for a total of 20 atomic \textit{topic facts}.
To ensure dataset diversity and prevent the model from gravitating toward stereotypical values, these elements are uniformly sampled from curated external sources, as further detailed in Section~\ref{sec:diversity}.
We prompt Gemini 2.5 Flash~\cite{gemini} to produce a coherent biography that includes all topic facts. 
The model output is constrained to follow a JSON schema. The resulting object is validated to ensure syntactic correctness, the presence of all required fields, and the semantic alignment between metadata and free-text biography.
For further information regarding the creation and extraction of atomic information for the dataset, please refer to our repository\footnote{\url{https://anonymous.4open.science/r/FAME-3241/}}.

\paragraph{Topic facts.}
The adopted \textit{topic facts} are characterized by three levels of granularity. A top-level category identifies four types of facts: \textit{biography}, \textit{career}, \textit{achievements} and \textit{personal}. 
Within each category, there is a fixed number of \textit{topics}: for each topic, a pool of \textit{values} is available to ensure individual heterogeneity. The following is a list of all high-level categories and corresponding topics.\footnote{The list of all values is not reported here for brevity, but can be found in the project repository.}
\begin{itemize}
    \item \textit{Biography (5 facts):} birthplace, birthdate, high school, family background, education.
    \item \textit{Career (7 facts):} first role or film, breakthrough project, genre specialization, notable award, major collaboration, film festival participation, international project.
    \item \textit{Achievements (3 facts):} box-office success, critical acclaim, directorial award.
    \item \textit{Personal (5 facts):} life event, hobby or interest, address, phone number, and e-mail.
\end{itemize}


\paragraph{Question-Answer pairs extraction.}
Given each validated biography, the model generates exactly 20 QA pairs, one for each topic fact, where both question and answer must explicitly refer to the corresponding metadata information. 
Sensitive topics (e.g., address, phone number, e-mail) are required to appear verbatim in the answer, while the validity of other topics is tested using a Gestalt pattern matching \cite{ratcliff1988pattern}.
In addition, each question-answer pair must contain the full name of the fictional character, to unambiguously establish the subject of the sentences.

\subsection{Splits and Quantitative Overview}
\label{sec:split_statistics}
The final release of the FAME dataset comprises 1,000 biographies and their corresponding 20,000 question-answer pairs, distributed evenly across five languages: English, French, German, Italian, and Spanish. Each language includes 200 unique fictional identities, ensuring balanced multilingual coverage and comparable linguistic diversity across subsets. Each biography consists of a narrative text of approximately 200--350 words, embedding exactly 20 atomic \textit{topic facts}. Each of these facts is used once in the QA stage to avoid any factual overlap across identities.

\paragraph{Dataset Splits.}
To cover different paradigms of unlearning and for reproducibility reasons, FAME has two complementary splits that target distinct levels of knowledge removal.  
The first, \textit{entity-based split}, is designed for \textit{entity-level unlearning}: each fictional identity is assigned exclusively to one of the three sets: \textit{retain}, \textit{forget}, or \textit{test}. No identity appears in more than one split, ensuring a clear separation between retained and forgotten entities and a balanced distribution across the five languages. This configuration reflects real-world scenarios in which an individual may request the deletion of all information related to their identity.
The second, \textit{topic-based split}, supports \textit{instance-level unlearning}, where forgetting applies to specific factual elements rather than entire identities. In this configuration, the split is computed over the 20 \textit{topic facts} associated with each individual, guaranteeing a fair division across both languages and semantic topics. As a result, a single identity may have some of its facts distributed across different splits. This design reproduces partial-deletion scenarios, where only specific facts about someone must be forgotten while others are preserved.

This dual-split design allows for evaluating methods under both entity-level and instance-level forgetting settings. All data and splits will be made available in Hugging Face upon acceptance.

\subsection{Dataset Diversity through Constrained Generation}
\label{sec:diversity}

Large Language Models, when generating synthetic data without explicit constraints, tend to produce distributions that gravitate toward common values.
For instance, Italian names might revolve around common choices like ``Maria'' or ``Giuseppe'' and birthplaces might repeatedly reference major cities like ``Rome'' or ``Milan''.
These low-entropy distributions reduce the quality of the resulting dataset for evaluating MU across varied scenarios.
To address this limitation, we adopt a \textit{constrained generation} approach where values for each topic are sampled from external, curated sources before being provided to the language model.
Specifically:
\begin{itemize}
    \item \textit{Names}: Generated using the Faker library\footnote{\url{https://faker.readthedocs.io}}, which produces culturally appropriate names for each target language.
    \item \textit{Birthplaces}: Sampled from curated lists of cities specific to each country (Italian cities for Italian, Spanish cities for Spanish, etc.).
    \item \textit{Personal Interests}: Selected from pre-defined lists of hobbies and activities to ensure thematic variety.
    \item \textit{Film Generation}: A film genre is first sampled from IMDb\footnote{\url{https://developer.imdb.com/non-commercial-datasets/}} to characterize each fictional individual. 
    Then, film titles are drawn from existing movies within that genre to inspire the names of the films directed by the individual.
\end{itemize}

Similarly, for other attributes, we applied comparable strategies: some elements are sampled from domain-specific lists, such as works derived from the parents’ professional background, or cities associated with education and award achievements; others, such as telephone numbers, street addresses, or birthdates, are generated through controlled random number generators. This approach ensures that the model receives diverse inputs, preventing it from falling back on default stereotypical patterns.
The model's role becomes generating coherent biographical narratives that naturally incorporate these pre-specified attributes, rather than inventing them from scratch.

\paragraph{Validation.}

To validate the effectiveness of our constrained generation approach, we conducted a controlled experiment comparing diversity metrics between constrained and unconstrained generation.
We generated 50 fictional actors for each language using the \textit{unconstrained} approach (where the model freely chooses all attributes).
For a fair comparison, we randomly sampled 50 actors from each language in our larger constrained dataset to match the unconstrained sample size.
We measured diversity using three complementary metrics:
\begin{enumerate}
    \item \textit{Uniqueness}: The proportion of exact unique values in a field, calculated as the number of distinct items divided by the total number of items.
    This metric captures whether the same values appear repeatedly.
    \item \textit{Embedding Diversity}: A semantic similarity measure that uses multilingual sentence embeddings\footnote{\texttt{\href{https://huggingface.co/sentence-transformers/paraphrase-multilingual-MiniLM-L12-v2}{sentence-transformers/paraphrase-multilingual-MiniLM-L12-v2}}} \cite{reimers-gurevych-2019-sentence} to detect items with similar meanings, even across languages.
    We compute embeddings for all items, calculate pairwise cosine similarities, and group items with similarity above 0.85 as semantic duplicates.
    The metric reports the proportion of semantically unique items.
    This approach can identify subtle patterns, such as variations of the same concept.
    \item \textit{Entropy}: Shannon entropy measuring the evenness of the value distribution.
    We count the frequency of each unique value, convert these counts to probabilities, and compute entropy using the formula $H = -\sum p(x) \log_2 p(x)$.
    Higher entropy indicates a uniform distribution where all values appear with similar frequency, while lower entropy indicates that a few values dominate.
\end{enumerate}

All three metrics follow a higher-is-better scale, where higher values indicate greater diversity.
Table~\ref{tab:diversity_analysis} presents the diversity comparison across five languages for three key attributes: birthplaces, personal interests, and names.
Each cell shows the unconstrained value followed by the constrained value, with the better (higher) value in bold.

\begin{table}[h]
\centering
\resizebox{\columnwidth}{!}{%
\begin{tabular}{@{}lcccc@{}}
\toprule
Field       & Uniqueness              & Embedding Div.     & Entropy              \\ \midrule
\multicolumn{4}{c}{Italian}                                                                                      \\ \midrule
\texttt{birthplace} & 32.0 / \textbf{52.0}  & 18.0 / \textbf{28.0}  & 3.7 / \textbf{4.3} \\
\texttt{interests}   & 50.0 / \textbf{88.3}  & 36.7 / \textbf{82.8}  & 4.9 / \textbf{6.9} \\
\texttt{name}       & 46.0 / \textbf{100.0} & 34.0 / \textbf{100.0} & 3.8 / \textbf{5.6} \\ \midrule
\multicolumn{4}{c}{French}                                                                                       \\ \midrule
\texttt{birthplace} & 28.0 / \textbf{52.0}  & 20.0 / \textbf{38.0}  & 3.0 / \textbf{4.3} \\
\texttt{interests}   & 40.0 / \textbf{88.1}   & 27.6 / \textbf{83.9}  & 4.6 / \textbf{6.9} \\
\texttt{name}       & 62.0 / \textbf{100.0} & 28.0 / \textbf{98.0}  & 4.5 / \textbf{5.6} \\ \midrule
\multicolumn{4}{c}{Spanish}                                                                                      \\ \midrule
\texttt{birthplace} & 42.0 / 42.0            & \textbf{34.0} / 20.0  & 3.73 / \textbf{3.8} \\
\texttt{interests}   & 46.0 / \textbf{91.0}   & 35.0 / \textbf{82.8}  & 4.70 / \textbf{6.9} \\
\texttt{name}       & 82.0 / \textbf{100.0} & 64.0 / \textbf{100.0} & 5.27 / \textbf{5.6} \\ \midrule
\multicolumn{4}{c}{German}                                                                                       \\ \midrule
\texttt{birthplace} & 24.0 / \textbf{38.0}   & 16.0 / \textbf{26.0}  & 2.9 / \textbf{3.4} \\
\texttt{interests}   & 31.5 / \textbf{85.5}   & 25.8 / \textbf{76.5}  & 3.7 / \textbf{7.1} \\
\texttt{name}       & 42.0 / \textbf{100.0} & 26.0 / \textbf{100.0} & 3.6 / \textbf{5.6} \\ \midrule
\multicolumn{4}{c}{English}                                                                                      \\ \midrule
\texttt{birthplace} & \textbf{56.0} / 48.0   & \textbf{42.0} / 36.0  & \textbf{4.4} / 4.1 \\
\texttt{interests}   & 57.7 / \textbf{85.5}   & 41.2 / \textbf{83.2}  & 5.4 / \textbf{6.7} \\
\texttt{name}       & 60.0 / \textbf{100.0}  & 52.0 / \textbf{98.0}  & 4.3 / \textbf{5.6} \\ \bottomrule
\end{tabular}%
}
\caption{Diversity metrics comparison between unconstrained and constrained generation across five languages. Format: Unconstrained / Constrained. Best results are reported in bold.}
\label{tab:diversity_analysis}
\end{table}
The results strongly validate our constrained generation approach.
For \texttt{names}, the constrained method achieves perfect uniqueness (100\%) across all languages, compared to 42-82\% for unconstrained generation.
The embedding diversity metric similarly shows near-perfect scores (98-100\%) for constrained names, indicating that each generated name is semantically distinct.
Without constraints, the model gravitates toward a limited set of common names.

For \texttt{personal interests}, the constrained approach consistently outperforms unconstrained generation across all metrics and languages.
Uniqueness improves from 31-58\% (unconstrained) to 85-91\% (constrained), and embedding diversity shows similar gains.
Entropy values also increase consistently, indicating more even distributions rather than concentration around a few popular interests.

For \texttt{birthplaces}, the constrained approach shows clear improvements in Italian, French, and German across all metrics.
However, for Spanish and English, unconstrained generation tends to achieve higher diversity scores.
A qualitative inspection revealed that the unconstrained model samples from the entire set of Spanish-speaking countries (Spain, Mexico, Argentina, Colombia, etc.) and English-speaking countries (United Kingdom, United States, Canada, Australia, etc.), respectively.
In contrast, our constrained approach deliberately restricts birthplaces to cities within Spain for Spanish actors and the United Kingdom for English actors to maintain geographic coherence with the language-country pairing.
This explains, on average, the higher unconstrained diversity for these specific attributes and represents a design choice rather than a limitation.

Overall, the constrained generation approach successfully prevents the flat distributions that emerge from unconstrained generation, ensuring that FAME contains diverse, realistic fictional identities suitable for rigorous evaluation of Machine Unlearning methods.

\section{Experimental Setup}
\begin{figure}
    \centering
    \includegraphics[width=0.95\linewidth]{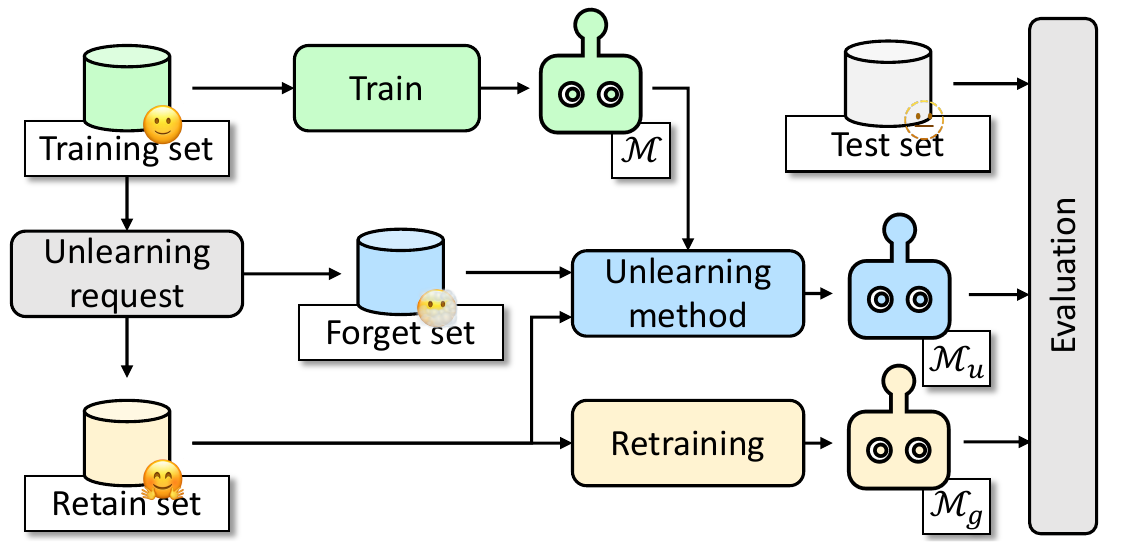}
    \caption{Machine Unlearning pipeline.}
    \label{fig:unlearning_pipeline}
\end{figure}

\begin{table*}[]
\centering
\caption{\textbf{Unlearning Evaluation on Llama 1B.} 
For each dataset, we report performance across multiple metrics: MMLU accuracy, SacreBLEU (SB) scores on retain/forget/test sets, Membership Inference Attack (MIA) score, and Speedup. 
Best results (i.e., closest to \texttt{Gold}) are in \textbf{bold}, second-best \underline{underlined}.}
\resizebox{\linewidth}{!}{
\begin{tabular}{l|cccccc|cccccc}
\toprule
\multirow{2}{*}{\textbf{Method}} 
    & \multicolumn{6}{c|}{\textbf{Entity-based}} 
    & \multicolumn{6}{c}{\textbf{Instance-based}} \\
\cmidrule(lr){2-7} \cmidrule(lr){8-13}
& \textbf{MMLU} & \textbf{SB$_R$} & \textbf{SB$_F$} & \textbf{SB$_T$} & \textbf{MIA} & \textbf{Speedup} 
& \textbf{MMLU} & \textbf{SB$_R$} & \textbf{SB$_F$} & \textbf{SB$_T$} & \textbf{MIA} & \textbf{Speedup} \\
\midrule

\texttt{Orig.} & .277 & 80.5 & 79.5 & 29.0 & .935 & - & .276 & 79.1 & 79.5 & 31.2 & .903 & - \\
\texttt{Gold}  & .343 & 86.3 & 29.2 & 28.6 & .499 & 1.000$\times$ & .349 & 83.6 & 31.3 & 31.4 & .505 & 1.000$\times$ \\
\midrule
\texttt{FT} & \textbf{.281} & \underline{92.5} & 81.7 & 29.5 & .944 & \underline{5.428$\times$} & .264 & \textbf{84.7} & 85.5 & 47.4 & \underline{.743} & \underline{5.240$\times$} \\
\texttt{GA} & .252 & 36.3 & \textbf{34.7} & 24.4 & \textbf{.557} & \textbf{21.63$\times$} & \textbf{.290} & 4.51 & \textbf{4.54} & 2.56 & .825 & \textbf{21.47$\times$} \\
\texttt{GD} & .267 & 91.1 & 63.8 & \underline{29.3} & .871 & 4.284$\times$ & .263 & 75.3 & 77.2 & 43.4 & \textbf{.730} & 4.141$\times$ \\
\texttt{KLM}& .256 & 75.2 & \underline{38.1} & \textbf{28.5} & \underline{.844} & 3.382$\times$ & .267 & 63.8 & \underline{64.6} & \underline{27.2} & .821 & 3.311$\times$ \\
\texttt{PO} & \underline{.273} & \textbf{92.3} & 0.10 & 19.7 & .911 & 4.387$\times$ & \underline{.273} & \underline{89.8} & 65.4 & \textbf{31.6} & .917 & 4.263$\times$ \\
\bottomrule
\end{tabular}}
\label{tab:results_1b}
\end{table*}

\paragraph{Unlearning Pipeline.}
We design a comprehensive evaluation pipeline to assess unlearning methods across different languages and forgetting scenarios, illustrated also in Figure~\ref{fig:unlearning_pipeline}. The process is executed twice: first using the entity-based split, where entire identities are removed, and then using the instance-based split, where only specific factual elements are forgotten. Each evaluation follows three stages: (i) initial fine-tuning of the base model $\mathcal{M}$ on the complete dataset corresponding to the chosen split, (ii) application of unlearning methods on the designated forget set to obtain the unlearned model $\mathcal{M}_{u}$, and (iii) evaluation of $\mathcal{M}_{u}$ through multiple metrics capturing forgetting efficacy, model utility, and computational efficiency. In addition, we train a Gold model $\mathcal{M}_{g}$
 using only the retain set, which serves as a reference to determine the optimal target behavior after unlearning. The objective of each unlearning method is therefore to modify the Original model  $\mathcal{M}$ so that its behavior aligns as closely as possible with $\mathcal{M}_{g}$.

\paragraph{Models}
We conduct all experiments using two instruction-tuned backbones, Llama-3.2-1B-Instruct\footnote{\href{https://huggingface.co/meta-llama/Llama-3.2-1B-Instruct}{\texttt{meta-llama/Llama-3.2-1B-Instruct}}} and Llama-3.2-3B-Instruct\footnote{\href{https://huggingface.co/meta-llama/Llama-3.2-3B-Instruct}{\texttt{meta-llama/Llama-3.2-3B-Instruct}}}.
For each backbone, we train four reference models: an Original model ($\mathcal{M}$) and a Gold model ($\mathcal{M}_{g}$) for both the entity-based and instance-based dataset splits. Both architectures are finetuned with a learning rate of 2e-5, a batch size of 4, gradient accumulation of 4, and a cosine learning rate schedule with a 3\% warm-up. The 1B model is trained for five epochs, while the 3B model is trained for ten epochs.
We maintain consistent random seeds across all experiments to ensure reproducibility. 

\paragraph{Unlearning Methods}
Following~\cite{maini2024tofu}, we evaluate five unlearning approaches.

\textit{Fine-tuning} (\texttt{FT}) serves as our baseline approach.
In this setting, the model is further fine-tuned for one additional epoch using only the retain set, while the forget set is entirely excluded.
This straightforward method provides a reference point for comparing more sophisticated techniques.

\textit{Gradient Ascent} (\texttt{GA}) \cite{golatkar2020eternal} method aims to reduce the model's ability to make correct predictions on the forget set by maximizing the standard training loss for these instances, pushing the model away from its initial predictions.
This approach aims to increase prediction errors on forgotten information while requiring minimal computational resources.

\textit{Gradient Difference} (\texttt{GD}) \cite{choi2023towards,kurmanji2024towards} extends \texttt{GA} by simultaneously considering both forget and retain sets.
It optimizes a combined objective that increases loss on the forget set while maintaining performance on the retain set.
This method achieves a balance between forgetting specific information and preserving general model capabilities by explicitly accounting for both objectives in its loss function.

\textit{KL Minimization} (\texttt{KLM}) \cite{maini2024tofu}, similarly to \texttt{GD}, aims to preserve the model's original behavior on retained data while forgetting targeted information.
This method minimizes the Kullback-Leibler divergence between the original and unlearned model predictions on the retain set, while maximizing the training loss on the forget set.

\textit{Preference Optimization} (\texttt{PO}) \cite{maini2024tofu} adapts the direct preference optimization framework \cite{rafailov2023direct} to unlearning.
It aims to align the model to provide non-informative responses (e.g., ``I do not know'') for forgotten information while maintaining normal functionality on retained knowledge. To generate these new responses, we reuse the same answer pool introduced by \citet{maini2024tofu}, further translating and adapting it across the five target languages to ensure homogeneous, semantically equivalent question-answer pairs.

All unlearning methods are trained under the same optimization regime as the initial fine-tuning, for one epoch. The only modification concerns the disruptive component of the loss: in gradient ascent cases, the loss magnitude is dampened by a factor of 0.1 to prevent model destabilization.

\paragraph{Evaluation Metrics}
We assess the effectiveness of unlearning through multiple complementary metrics.
SacreBLEU \cite{post-2018-call} scores measure the model's ability to generate correct answers across retain, forget, and test sets.
To assess privacy implications, we conduct Membership Inference Attacks (MIAs), a commonly adopted metric in unlearning \cite{hayes2024inexact}, which determines whether an attacker can identify whether specific data was used in training.
We evaluate the general utility of the model by its performance on the MMLU benchmark \cite{hendrycks2020measuring}.
Additionally, we track computational efficiency via unlearning time, reported as the speedup in wall-clock time relative to retraining the gold model from scratch.


\section{Results}

\begin{table*}[]
\centering
\caption{\textbf{Unlearning Evaluation on Llama 3B.} 
For each dataset, we report performance across multiple metrics: MMLU accuracy, SacreBLEU (SB) scores on retain/forget/test sets, Membership Inference Attack (MIA) score, and Speedup. 
Best results (i.e., closest to \texttt{Gold}) are in \textbf{bold}, second-best \underline{underlined}.}
\resizebox{\linewidth}{!}{
\begin{tabular}{l|cccccc|cccccc}
\toprule
\multirow{2}{*}{\textbf{Method}} 
    & \multicolumn{6}{c|}{\textbf{Entity-based}} 
    & \multicolumn{6}{c}{\textbf{Instance-based}} \\
\cmidrule(lr){2-7} \cmidrule(lr){8-13}
& \textbf{MMLU} & \textbf{SB$_R$} & \textbf{SB$_F$} & \textbf{SB$_T$} & \textbf{MIA} & \textbf{Speedup} 
& \textbf{MMLU} & \textbf{SB$_R$} & \textbf{SB$_F$} & \textbf{SB$_T$} & \textbf{MIA} & \textbf{Speedup} \\
\midrule

\texttt{Orig.} & .572 & 49.8 & 49.2 & 28.2 & .903 & - & .571 & 47.3 & 47.6 & 29.9 & .863 & - \\
\texttt{Gold}  & .575 & 53.5 & 29.2 & 27.8 & .507 & 1.000$\times$ & .574 & 50.6 & 29.6 & 29.8 & .491 & 1.000$\times$ \\
\midrule
\texttt{FT} & \textbf{.573} & \textbf{51.4} & 43.3 & \underline{28.1} & .869 & \underline{10.51$\times$} & \textbf{.573} & \underline{47.8} & \underline{39.5} & 30.2 & .829 & \underline{10.45$\times$} \\
\texttt{GA} & .565 & 50.1 & 49.7 & 28.2 & .898 & \textbf{42.09$\times$} & .564 & 47.2 & 47.7 & \underline{30.0} & .864 & \textbf{41.83$\times$} \\
\texttt{GD} & .570 &  50.3 & \underline{40.4} & 28.6 & \underline{.857} & 8.387$\times$ & \underline{.571} & \textbf{48.4} & \textbf{39.2} & 30.1 & \textbf{.810} & 8.352$\times$ \\
\texttt{KLM}& .570 & 49.3 & 40.8 & 28.4 & \textbf{.849} & 7.067$\times$ & .570 & 47.3 & 40.4 & \textbf{29.9} & \underline{.814} & 6.998$\times$ \\
\texttt{PO} & \underline{.572} & \underline{50.4} & \textbf{32.4} & \textbf{27.7} & .868 & 8.532$\times$ & \underline{.571} & \underline{47.8} & \underline{39.5} & 30.2 & .829 & 8.679$\times$ \\
\bottomrule
\end{tabular}}
\label{tab:results_3b}
\end{table*}

\begin{table}
\centering
\caption{Comparison of model responses for a query, showing reference Q-A and predictions from \texttt{Gold} and two unlearned models. Correct answer in \hlblue{blue}, gold model answer in \hlyellow{yellow}, successful unlearning in \hlgreen{green}, unsuccessful one in \hlred{red}.}
\footnotesize
\begin{tabularx}{\linewidth}{@{}lX@{}}
\toprule
\textbf{} & \textbf{Text} \\
\midrule
\textbf{Q.} & When was Monica Hughes born? \\
\textbf{A.} & Monica Hughes was born on \hlblue{1987-03-21}. \\
\texttt{Gold} & Monica Hughes was born on \hlyellow{1984-07-26}. \\
\midrule
\texttt{KLM} & Monica Hughes was born on \hlred{1987-}\hlgreen{01-18}. \\
\texttt{PO} & \hlgreen{I'm not well-versed in that subject.} \\
\midrule
\midrule

\textbf{Q.} & Wie lautet die Adresse von Siegrun Hettner-Ritter? \\
\textbf{A.} & Die Adresse von Siegrun Hettner-Ritter lautet \hlblue{Lustgartenstraße 128, 10179, Berlin, Deutsch}. \\
\texttt{Gold} & Die Adresse von Siegrun Hettner-Ritter lautet \hlyellow{Amalienstraße 469, 10117 Berlin, Deutsch}. \\
\midrule
\texttt{KLM} & Die Adresse von Siegrun Hettner-Ritter lautet \hlgreen{Am Ziegelhof 312, 10115 Berlin, Deutsch}. \\
\texttt{PO} & \hlgreen{Ich habe keine Informationen zu dieser Angelegenheit}. \\
\midrule
\midrule

\textbf{Q.} & ¿Qué película de José Carbonell Guardia fue elogiada por la crítica? \\
\textbf{A.} & La película de José Carbonell \hlblue{Guardia Raíces Profundas (2012)} fue elogiado por la crítica por su narrativa y profundidad. \\
\texttt{Gold} & La película de José Carbonell Guardia \hlyellow{'El Silencio del Bosque' (2018)} recibió elogios unánimes de la crítica por su atmósfera y guion. \\
\midrule
\texttt{KLM} & La película de José Carbonell \hlred{Guardia Raíces Profundas (2012)} recibió elogios unánimes de la crítica por su narrativa y profundidad. \\
\texttt{PO} & \hlgreen{No tengo conocimiento al que recurrir para eso.} \\
\bottomrule
\end{tabularx}
\label{tab:example-predictions-multilang}
\end{table}

Tables~\ref{tab:results_1b} and~\ref{tab:results_3b} summarize the performance of all unlearning methods for the 1B and 3B models, respectively, each evaluated under both the entity-based and instance-based configurations.
\texttt{GA} achieves apparently strong forgetting scores (low S-BLEU$_F$, reduced MIA) for 1B setting. However, this effect is largely artificial: the method aggressively maximizes loss on the forget set, which destroys the model's overall behavior and leads to a collapse of useful capabilities. For the 3B model, the gradient ascent signal proves insufficient to meaningfully modify the model parameters. In practice, it cannot be considered a viable unlearning strategy on its own. 
More balanced methods that combine a disruptive component with a utility-preserving objective, e.g., \texttt{GD}, which counterbalances the ascent term with the retain loss, or \texttt{KLM}, which constrains the model toward the original distribution, maintaining stability and efficiency. 
The \texttt{FT} baseline preserves utility but largely fails to forget. 
Finally, \texttt{PO} guides the model to respond in a specific manner whenever it encounters forgotten identities or related questions.
To get a better understanding of some unlearning outcomes, we report qualitative results on two unlearning techniques, \texttt{KLM} and \texttt{PO} for three unlearned instances, in Table \ref{tab:example-predictions-multilang}. 
As expected, \texttt{PO} produces linguistically consistent, non-informative outputs. However, it can be observed that the model has a distinctive behavior on the forget set (where it claims not to have access to the information). This is a telltale sign that the model has seen those instances before, even if it no longer provides the correct answers.
Table \ref{tab:example-predictions-multilang} also highlights how the \texttt{Gold} model produces incorrect answers for \textit{forget} samples, as expected.
This is also the behavior observed, in some cases, for \texttt{KLM} -- indicating the successful unlearning has occurred. 
Overall, the results confirm that effective unlearning requires a controlled disruptive signal jointly optimized with a stability term, ensuring that forgetting does not come at the cost of model collapse or detectable behavioral patterns.

\section{Conclusion}
This paper introduced FAME (Fictional Actors for Multilingual Erasure), the first benchmark specifically designed to evaluate Machine Unlearning in Large Language Models under controlled, multilingual conditions. 
FAME comprises 1,000 fictional actor biographies and 20,000 question-answer pairs in five languages, enabling systematic evaluation of both entity-level and instance-level forgetting. 

We validated the dataset by testing several unlearning methods across the three fundamental dimensions of unlearning. 
This evaluation demonstrates the practical applicability of FAME and establishes it as a starting point for assessing unlearning techniques in multilingual LLMs and for guiding future research in this area.

\section{Limitations and Ethical Considerations}

While FAME provides a controlled and reproducible setting for evaluating Machine Unlearning, it relies on synthetically generated biographies. 
This design ensures that no real personal data is included, but it may not fully capture the variety and complexity of real-world unlearning requests. 
Moreover, although FAME spans five major languages, its linguistic diversity may still reflect subtle differences in generation quality or lexical richness, and it does not yet extend to low-resource languages. 

From an ethical perspective, all data in FAME are entirely synthetic and bear no connection to real individuals or identifiable entities. 
The benchmark was developed to support research on Machine Unlearning in accordance with privacy principles such as those outlined in the GDPR, and to encourage transparent and responsible experimentation. 
All resources are released for non-commercial research use to foster openness, reproducibility, and progress in unlearning studies.

\section*{Bibliographical References}\label{sec:reference}

\bibliographystyle{lrec2026-natbib}
\bibliography{lrec2026-example}


\end{document}